\pdfoutput=1

\documentclass[11pt]{article}

\usepackage{emnlp2021}

\usepackage{times}
\usepackage{latexsym}

\usepackage[T1]{fontenc}

\usepackage[utf8]{inputenc}

\usepackage{microtype}

\usepackage{multirow}
\usepackage{bm}
\usepackage{amsmath}
\usepackage{mathtools}
\usepackage{graphicx}
\usepackage{cuted}
\usepackage{tikz}
\usepackage{caption}
\usepackage{subcaption}
\usepackage{xurl}

\newcommand{\argmin}{\mathop{\mbox{argmin}}}

\DeclarePairedDelimiter\floor{\lfloor}{\rfloor}

\title{Stream-level Latency Evaluation for Simultaneous Machine Translation}

\author{Javier Iranzo-S\'{a}nchez \and  Jorge Civera \and   Alfons Juan \\
Machine Learning and Language Processing Group\\
Valencian Research Institute for Artificial Intelligence\\
Universitat Polit\`{e}cnica de Val\`{e}ncia\\
Cam\'{\i} de Vera s/n, 46022 Val\`{e}ncia, Spain \\
  \texttt{ \{jairsan,jorcisai,ajuanci\}@vrain.upv.es  }\\}
\begin{document}
\maketitle
\begin{abstract}
Simultaneous machine translation has recently gained traction
thanks to significant quality improvements and the advent of 
streaming applications. Simultaneous translation systems need 
to find a trade-off between translation quality and response time, 
and with this purpose multiple latency measures have been proposed.
However, latency evaluations for
simultaneous translation are estimated at the sentence 
level, not taking into account the sequential nature of a streaming 
scenario. Indeed, these sentence-level latency measures 
are not well suited for continuous stream translation, resulting 
in figures that are not coherent with the simultaneous translation policy 
of the system being assessed. This work proposes a stream-level 
adaptation of the current latency measures based on a 
re-segmentation approach applied to the output translation, that is 
successfully evaluated on streaming conditions for a reference IWSLT task. 
\end{abstract}

\section{Introduction}

Simultaneous speech translation systems just started to become
available~\cite{Bahar2020,Elbayad2020b,Han2020,Pham2020} thanks 
to recent developments in streaming automatic speech recognition 
and simultaneous machine translation.
These systems seamlessly translate a continuous audio stream 
under real-time latency constraints. 
However, current translation latency evaluations~\cite{Ansari2020} 
are still performed at the sentence-level based on the conventional 
measures, 
Average Proportion (AP)~\citep{DBLP:journals/corr/ChoE16}, Average
Lagging (AL)~\citep{Ma2019} and Differentiable
Average Lagging (DAL)~\citep{Cherry2019}. These measures compute 
the translation latency for each sentence independently without taking 
into account possible interactions that lead to accumulated delays
in a real-world streaming scenario. Additionally, the current measures 
cannot be used by systems that do not use explicit sentence-level
segmentation~\cite{schneider-waibel-2020-towards}.

In this work, we first revisit the conventional translation latency 
measures in Section~\ref{sec:related} to motivate their adaptation to 
the streaming scenario in Section~\ref{sec:stream}. Then, these adapted 
latency measures are computed and reported on an IWSLT task in Section~\ref{sec:exps}. Finally, conclusions and future work are presented in 
Section~\ref{sec:con}.

\section{Related work}
\label{sec:related}

Current latency measures for simultaneous translation 
can be characterised as a
normalisation of the number of read-write word operations required
to generate a translation $\bm{y}$ from a source sentence $\bm{x}$
\begin{equation}
\label{eqn:L}
L(\bm{x},\bm{y}) = \frac{1}{Z(\bm{x},\bm{y})} \sum_{i} C_i(\bm{x},\bm{y})
\end{equation}
with $Z$ being a normalisation function, $i$ an index over the 
target positions and $C_i$ a cost function for each target position $i$.
Depending on the latency measure, $C_i$ is defined as
\begin{equation}
\label{eqn:C}
  C_i(\bm{x},\bm{y}) = 
  \begin{cases}
    g(i) & \text{AP}\\
    g(i) - \frac{i-1}{\gamma} & \text{AL}\\
    g'(i) - \frac{i-1}{\gamma} & \text{DAL}
  \end{cases} 
\end{equation}
with 
\begin{equation}
g'(i) = \max
  \begin{cases}
    g(i) \\
    g'(i-1) + \frac{1}{\gamma}
  \end{cases}
\end{equation}
where $g(i)$ is the number of source tokens read when a token is
written at position $i$ and $\gamma$ is target-to-source length ratio $\frac{|\bm{y}|}{|\bm{x}|}$. Note that the AP cost function 
considers the absolute number of source tokens that
has been read to output the $i$-th word, while AL and DAL cost
functions account for the number of source words the model lags behind
a wait-0 oracle. This oracle simply accumulates a uniform distribution of 
source words over target positions according to the ratio $\frac{1}{\gamma}$.
In the case of DAL, the recurrent definition of $g'(i)$ 
guarantees that the most expensive read-write operation is considered.

On the other hand, the normalisation function $Z$ depends on the measure
according to
\begin{equation}
\label{eqn:Z}
  Z(\bm{x},\bm{y}) = 
  \left\{ \!\!\! \begin{array}{ll}
    |\bm{x}|\cdot|\bm{y}| & \text{AP}\\
    \argmin\limits_{i:g(i)=|\bm{x}|} \; i & \text{AL}\\
    |\bm{y}| & \text{DAL}
  \end{array} \right. 
\end{equation}

The term in AP normalises the sum over the target sentence of absolute
source tokens, while AL and DAL does over the number of target
positions, which in the case of AL is limited to those target
positions reading new source tokens.  Indeed, the normalization term of AL
is referred to as $\tau$. The sentence-level latency measures just described are reported as an
average value over an evaluation set of multiple sentence pairs, each one
evaluated independently from the others.

However, the latency evaluation of a continuous paired stream of
sentences has not received much attention, with the exception of the
strategy proposed by~\cite{schneider-waibel-2020-towards}. 
This evaluation strategy considers the straightforward approach of
concatenating all sentences into a single source/target pair in order 
to compute the corresponding latency measure.  
Next section outlines some drawbacks of this strategy
(hereafter \textit{Concat-1}) to motivate the discussion on how 
the current sentence-level latency measures could be adapted
to the streaming scenario.

\section{Stream-level evaluation}
\label{sec:stream}

Let us consider the translation of a stream of two sentences,
the first sentence has two input and two output tokens, 
while the second 
one has two input and four output tokens with ratios $\gamma_1=1$ and $\gamma_2=2$, respectively. The translation
process is performed with a sentence-based wait-k system with catch-up 
characterised by a function $g(i)= \floor{k + \frac{i-1}{\gamma}}$ with $k=1$. 

Table~\ref{tab:example1} compares the computation of the latency measures 
for the Concat-1 strategy (top) with the conventional strategy that 
considers independent sentences (bottom). 
Note that the translation process has only been carried out once, 
but both strategies are just interpreting the results differently as 
first denoted by their $i$ and $g(i)$ values.
The wait-0 oracle $\frac{i-1}{\gamma}$ of Concat-1, with 
a single global $\gamma=\frac{3}{2}$ underestimates 
the actual writing rate, and the system accumulates more delay than in 
the evaluation strategy of independent sentences, which uses a sentence-level 
estimation for $\gamma$.

\newcommand{\STAB}[1]{\begin{tabular}{@{}c@{}}#1\end{tabular}}
\begin{table}
\centering
 \caption{Comparison of the latency metric computation between 
 the Concat-1 (top) and the conventional sentence-level (bottom) 
 strategy when using a wait-1 system.\label{tab:example1}}
\small 
	\begin{tabular}{ccc||l@{~~}l@{~~}|l@{~~}l@{~~}l@{~~}l|c}
\multirow{6}{*}{\STAB{\rotatebox[origin=c]{90}{Concat-1}}} &&	       & \multicolumn{6}{c|}{} & $L$\\\hline
&&	$i$    & 1 & 2   & 3   & 4   & 5   & 6   & \\
&&	$g(i)$ & 1 & 2   & 3   & 3   & 4   & 4   & \\
&&	$\frac{i-1}{\gamma}$ & 0 & $0.6$ & $1.3$ & 2.0 & $2.6$ & $3.3$ &\\
& \multirow{3}{*}{$C_i$} &	AP     & 1 & 2   & 3   & 3   & 4   & 4   & 
	$0.7$\\
&&	AL     & 1 & 1.3 & 1.6 & 1   & 1.3 & -   & 
	$1.2$\\
&&	DAL    & 1 & 1.3 & 1.6 & 1.6 & 1.6 & 1.6 & 
	$1.5$\\
	\hline
\multirow{6}{*}{\STAB{\rotatebox[origin=c]{90}{Ind. Sent.}}} &&    $i$    & 1 & 2 & 1 & 2 & 3 & 4 &\\
&&	$g(i)$ & 1 & 2 & 1 & 1 & 2 & 2 &\\
&&	$\frac{i-1}{\gamma}$ & 0.0 & 1.0 & 0.0 & 0.5 & 1.0 & 1.5 &\\ 
& \multirow{3}{*}{$C_i$} &	AP     & 1 & 2 & 1 & 1 & 2 & 2 &  
	$0.8$\\
&&	AL     & 1 & 1 & 1 & 0.5 & 1 & - & 
	$0.9$\\
&&	DAL    & 1 & 1 & 1 & 1   & 1 & 1 & 
	$1.0$\\
	\end{tabular}

\end{table}

These differences in results are magnified when computing latencies on a real streaming evaluation set.
On the one hand, AL and
DAL tend to obtain scores that do not reflect the real behaviour of the
system when using a Concat-1 strategy with a single global $\gamma$, since
the source-target length ratio
varies wildly between different sentences. Therefore, the wait-0 oracle will 
sometimes overestimate
the actual writing rate, and sometimes it will underestimate it. Moreover, the definition of
DAL keeps the system from recovering from previously incurred delays, 
and therefore, every time the writing rate is underestimated, the system falls further and further behind the wait-0 oracle. 
On the other hand, AP turns out to be little informative when the stream 
is long enough, since AP always tends to be 0.5 because the
delay incurred by a system with a reasonable $k$ is always negligible
compared with the total source length.

The accuracy of AL and DAL could be improved 
if sentence-level estimations for $\gamma$ would be available somehow
in a streaming scenario. With the availability of these estimations 
in mind, we formulate 
a streaming version of the cost functions in Eq.~\ref{eqn:C} based on 
a global $G(i)$ function, which returns the number of source tokens (including those from previous sentences) that have been read as in the Concat-1 strategy: 
 
\begin{equation}
\label{eqn:sC}
  C_{i}(\bm{x}_n,\bm{y}_n) = 
  \begin{cases}  
    g_n(i) & \text{AP}\\
    g_n(i) - \frac{i-1}{\gamma_n} & \text{AL}\\
    g_n'(i) - \frac{i-1}{\gamma_n} & \text{DAL}\\
  \end{cases} 
\end{equation}

with $g_n'(i)$ defined as
\begin{equation}
\max 
\begin{cases}
g_n(i)\\
\begin{cases}
g_{n-1}'(|\bm{x}_{n-1}|) + \frac{1}{\gamma_{n-1}} & i=1\\
g_n'(i-1) + \frac{1}{\gamma_{n}} & i>1
\end{cases} 
\end{cases} 
\end{equation}

where $g_n(i)=G(i+|\bm{y}_1^{n-1}|)-|\bm{x}_1^{n-1}|$. Thus, the global delay is converted to a local representation so that it can be compared with the local sentence oracle.

Table~\ref{tab:example_global} shows the computation of the stream-level latency measures as proposed in Eq.~\ref{eqn:sC} for the same example 
calculated in Table~\ref{tab:example1}. As observed, 
unlike with the Concat-1 strategy, we obtain the same results as in the conventional sentence-level estimation, while at the same time we keep the property that previous delays affect future sentences by basing our computations on the global delay $G(i).$

\begin{table}
\centering
 \caption{Estimation of stream-level latencies measures on the 
 same example proposed in Table~\ref{tab:example1}. \label{tab:example_global}}
\small 
	\begin{tabular}{rr||l@{~~}l@{~~}|l@{~~}l@{~~}l@{~~}l|c}
&	       & \multicolumn{6}{c|}{} & $L$\\\hline
&    	$i$    & 1 & 2 & 1 & 2 & 3 & 4 &\\
&	$G(i+|\bm{y}_1^{n-1}|)$ & 1 & 2   & 3   & 3   & 4   & 4   & \\
&	$g_n(i)$ & 1 & 2 & 1 & 1 & 2 & 2 &\\
&$\frac{i-1}{\gamma_n}$ & 0.0 & 1.0 & 0.0 & 0.5 & 1.0 & 1.5 &\\ 
\multirow{3}{*}{$C_i$} &	AP     & 1 & 2 & 1 & 1 & 2 & 2 &  
	$0.8$\\
&	AL     & 1 & 1 & 1 & 0.5 & 1 & - & 
	$0.9$\\
&	DAL    & 1 & 1 & 1 & 1   & 1 & 1 & 
	$1.0$\\
	\end{tabular}
\end{table}

If we use a segmentation-free model whose output is a single text
stream, stream-level latency measures can be still computed by 
re-segmenting the
output into sentence-like units (chunks). Formally, a segmenter takes an
input stream $Y$ and a set of reference sentences
to compute a re-segmentation $\bm{\hat{y}}_1^N$ of 
$Y$. Once the re-segmentation is obtained, stream-level latency 
measures are estimated by considering paired input-output segments 
$(\bm{x}_n,\bm{\hat{y}}_n)$. 
In our case, we re-segment by minimizing the
edit distance between the stream hypothesis and the reference
translations, analogously to the translation quality evaluation widely-used 
in speech translation~\cite{MatusovLBN05}. Likewise, we can resegment
the output to compute latency measures if our system uses a different
segmentation than the reference.

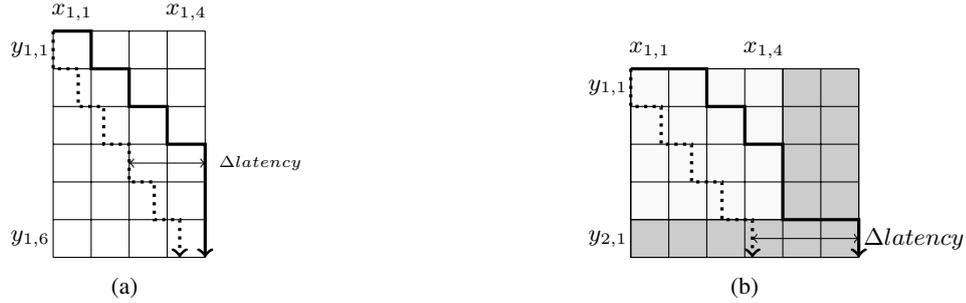
\begin{figure*}
\begin{subfigure}[b]{0.49\textwidth}        
\centering
    \begin{tikzpicture}[scale=0.5,yscale=-1]
        \draw[step=1.0,black,thin] (0.0,0.0) grid (4,6);
 \draw (-3,0);
    \draw (0.5,-0.5) node { \small $x_{1,1}$}; 
        \draw (3.5,-0.5) node { \small $x_{1,4}$}; 
    \draw (-0.6,0.5) node { \small $y_{1,1}$}; 
    \draw (-0.6,5.5) node { \small $y_{1,6}$}; 

\draw[->,very thick,black]  (0, 0) -> (1,0) -> (1,1) -> (2,1) -> (2,2) -> (3,2) -> (3,3) -> (4,3) -> (4,4) -> (4,5) -> (4,6);
\draw[->,very thick,black, dotted]  (0, 0) -> (0, 1) -> (0.66, 1) -> (0.66, 2) -> (1.33, 2) -> (1.33,3) -> (2, 3) -> (2, 4) -> (2.66, 4) -> (2.66, 5) -> (3.33, 5) -> (3.33, 6);

\draw[<->, thin] (2, 3.5 ) -> (4, 3.5);
    \draw (5.45, 3.5) node { \tiny $\Delta latency $}; 
 \end{tikzpicture}
         \caption{}
         \label{fig:latency_example_1}
     \end{subfigure}
     \hfill
\begin{subfigure}[b]{0.49\textwidth}        
\centering
    \begin{tikzpicture}[scale=0.5,yscale=-1]
 \draw (-3,0);
     \draw[fill=gray!40] (0.0,0.0) rectangle (6,5);
    \draw[fill=gray!05] (0.0,0.0) rectangle (4,4);
    
    \draw (0.5,-0.5) node { \small $x_{1,1}$}; 
    \draw (3.5,-0.5) node { \small $x_{1,4}$}; 

    \draw (-0.6,0.5) node { \small $y_{1,1}$}; 
    \draw (-0.6,4.5) node { \small $y_{2,1}$}; 
    
    \draw[step=1.0,black,thin] (0.0,0.0) grid (6,5);

    \draw[->,very thick,black]  (0, 0) -> (2,0) -> (2,1) -> (3,1) -> (3,2) -> (4,2) -> (4,3) -> (4,4) -> (6,4) -> (6,5);
    
    \draw[->,very thick,dotted]  (0, 0) -> (0,1) -> (0.8, 1) -> (0.8, 2) -> (1.6, 2) -> (1.6, 3) -> (2.4, 3) -> (2.4, 4) -> (3.2, 4) -> (3.2, 5);
    
    \draw[<->, thin] (3.2, 4.5 ) -> (6, 4.5);
    \draw (7.45, 4.5) node { \small $\Delta latency $}; 
 	\end{tikzpicture}

	\caption{}
         \label{fig:latency_example_1}
     \end{subfigure}
	\caption{The examples shown above illustrate how a model which follows a wait-$k$ policy can obtain AL/DAL values that differ from $k$. The bold lines show the
behaviour of the model, the dotted lines show the oracle policy. Left: writing rate error with $k=1$; the model uses $\hat \gamma=1$, but the actual value is $\gamma=1.5$. Right: segmentation error with  $k=2$; the first translated word of the second sentence is wrongly assigned to the first sentence during resegmentation, i.e. $\hat y_1=(y_{1,1},y_{1,2}, y_{1,3}, y_{1,4},y_{2,1})$. }
         \label{fig:latency_example}
\end{figure*}

Moreover, stream-level AL and DAL measures computed for a wait-$k$ system
are coherently close to $k$ with two caveats. 
First, there can be deviation
from the theoretical value of $k$ due to a inaccurate estimation of the
writing rate. Given that the wait-$k$ policy uses a fixed $\gamma$,
there will be some sentences in which this results in lower or higher
writing rates than desirable. This is a feature inherent to the fixed
policy itself. Second, a deviation could also occur due to
re-segmentation errors. For instance, a word that is part of
the translation of the $n$-th segment can be wrongly included into 
the previous $n-1$-th segment causing an increase of
the latency. Both sources of latency are illustrated in Figure \ref{fig:latency_example}.

These two caveats given the definition of DAL
imply that a system can never recover from previous delays, 
which might be an
acceptable solution when computing latency measures at the sentence level, 
but it seems too strict and unrealistic when computing latency measures for streams comprising
tens of thousands of words. To alleviate this problem, we propose to multiply the cost of a write operation $\frac{1}{\gamma_n}$ in $g_n'(i)$ by a scaling factor $s\in[0,1]$. In practice, for values of $s$ close to $1$,  this
means that the write operation costs slightly less for the real system 
than for the oracle. We believe this is an acceptable practical solution given that
there are many ways that this could be achieved in real-world tasks, such as rendering
subtitles slightly faster or, in the case of cascade
speech-to-speech, slight reducing the duration of TTS segments or increasing 
the playback speed. Finally, the scaling factor $s$ can be also understood 
as a hyperparameter that bridges the gap between AL ($s=0$) and DAL ($s=1$) 
and it can be adjusted depending on the actual writing cost of the translation task.

\section{Experiments}
\label{sec:exps}

The  stream-level latency measures proposed 
in Section~\ref{sec:stream} are now computed and evaluated on 
the IWSLT 2010 German-English dev set~\cite{Paul2010}. To simulate 
a streaming scenario, all source sentences are concatenated into a 
single input stream. Then, they are segmented into sentences
and translated with a wait-$k$ fixed policy. As a result, it is expected 
that a well-behaved latency measure should rank the systems
by increasing order of $k$.

Our streaming simultaneous translation system 
is based on a direct segmentation (DS) model~\cite{Iranzo2020b} followed 
by a Transformer BASE model~\cite{VaswaniSPUJGKP17}
trained with the multi-$k$ approach~\cite{Elbayad2020}.
The DS model was trained on TED talks~\cite{Cettolo2012}
with a future window of length 0 and history size of 10,
while the translation model was trained on the IWSLT 2020
German-English data~\cite{Ansari2020}. This system,
which we will refer to as \emph{Real}, uses catch-up
with $\gamma=1.24$. In addition 
to the Real system, three experimental setups based on different oracles are considered:
\begin{itemize}
 \item \emph{In. Seg.}: The input segmentation provided by the DS model is 
 replaced by the reference segmentation to gauge segmentation errors.
 \item \emph{Out. Seg.}: The reference segmentation is used to link each translation
 with its corresponding source sentence, therefore avoiding the need of re-segmentation by minimum edit distance.
 \item \emph{Policy}: The translation model is replaced by an oracle model 
 that outputs the reference translation with the appropriate writing rate
 for each sentence to account for errors due to a global $\gamma$. 
\end{itemize}

AL (Table \ref{tab:concat_1_al}) and DAL (Table \ref{tab:concat_1_dal}) have been computed using the Concat-1 approach, to serve as a baseline
for the developed measures. These results confirm the problems of the Concat-1 approach, which have been identified 
and discussed on Section \ref{sec:stream}. AP results have been excluded from the tables, as 
no matter which setup is used, the computed AP is always 0.5. Likewise, the obtained AL and DAL values offer little insight
about the latency behaviour of the model. These results are not only uninterpretable, but they also alter the ranking of
the models. This could be specially worrisome if the Concat-1 approach was used to compare systems with adaptative
policies that lack a explicit latency control such as $k$, as it might be harder to detect wheter the incoherent results are
due to the adaptative policy or the latency measure itself. The only setup which returns the correct ranking is the one using 
the In. Seg. and Policy Oracles, but the latency results do not reflect the real behaviour
of the model. The full AL and DAL results, for values up to $k=10$ are reported in the appendix.

\addtolength{\tabcolsep}{-2pt}    
\begin{table}[t]
\centering \caption{Stream-level AL as a function of $k$, computed using the Concat-1 approach on the IWSLT 2010 German-English dev set. \label{tab:concat_1_al}}
\begin{tabular}{l||rrrrr}
System & 1 & 2 & 3 & 4 & 5  \\ \hline
Real & -9.7 & -12.0 & -45.2 & -23.7 & -8.5  \\
~ +In. Seg.  & -42.9 & -29.0 & 17.4 & -10.1 & 25.5  \\
~~  + Policy  & 14.2 & 15.1 & 16.0 & 16.8 & 17.6 \\
\end{tabular}
\end{table}
\addtolength{\tabcolsep}{2pt}    

\begin{table}[h]
\centering \caption{Stream-level DAL as a function of $k$, computed using the Concat-1 approach on the IWSLT 2010 German-English dev set. \label{tab:concat_1_dal}}
\begin{tabular}{l||rrrrr}
System & 1 & 2 & 3 & 4 & 5  \\ \hline
Real &  15.0 & 11.0 & 17.4 & 11.3 & 20.3   \\
~ +In. Seg.  & 4.5 & 8.5 & 37.1 & 24.6 &  52.3  \\
~~  + Policy  & 85.8 & 86.7 & 87.7 & 88.7 & 89.7 \\
\end{tabular}
\end{table}

\begin{figure*}[h!] 
\centering
\includegraphics[width=.36\textwidth]{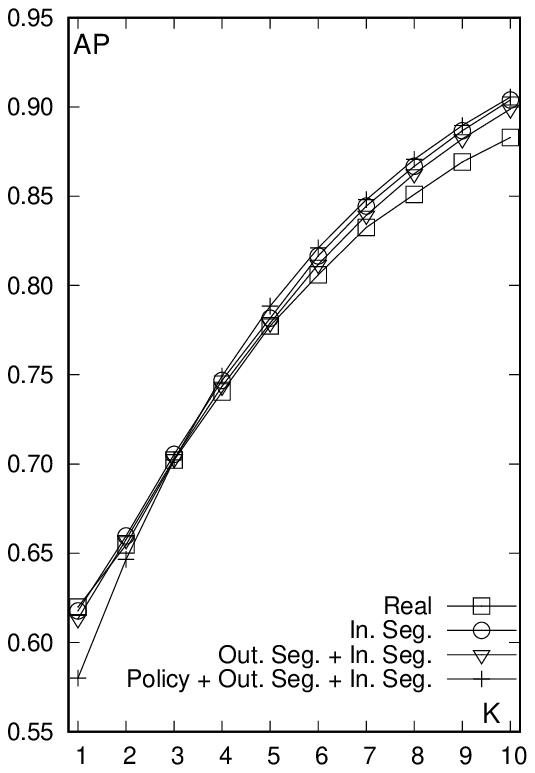}\hspace*{-5mm}
\includegraphics[width=.36\textwidth]{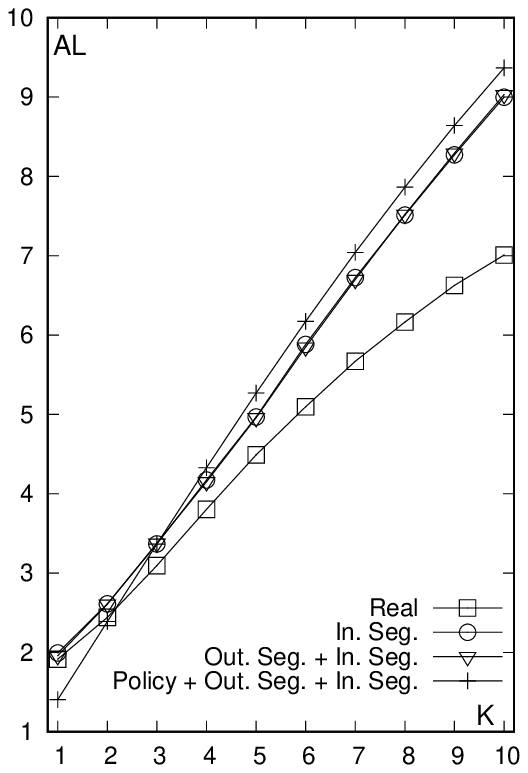}\hspace*{-5mm}
\includegraphics[width=.36\textwidth]{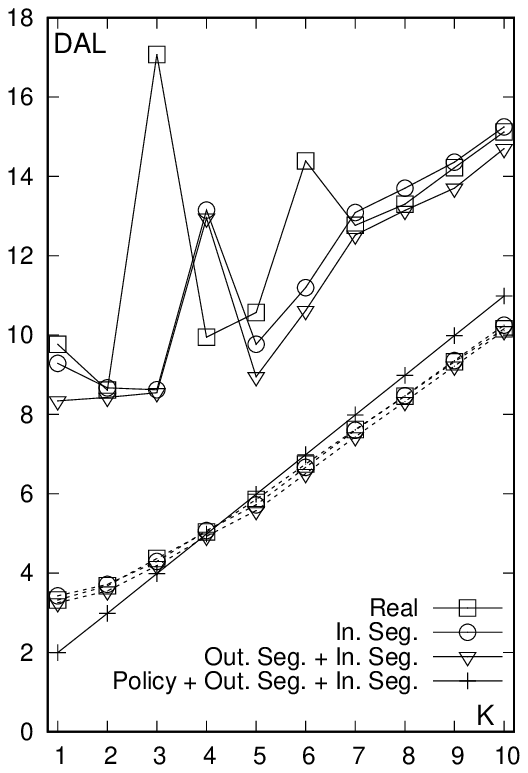}

\caption{
Stream-level AP (left), AL (center) and DAL (right) with $s=1.0$ 
and $s=0.95$ (dashed lines) as a function of $k$ 
in the multi-$k$ approach for four experimental setups
on the IWSLT 2010 German-English dev set.\label{fig:stream}}
\end{figure*}

Now that we have experimentally shown that the Concat-1 approach is unable to properly compute latencies, we move onto computing the stream-adapted version of the latency measures. The computation of stream-level AP (left), AL (center) and, DAL (right) 
with $s=1.0$ and $s=0.95$ (dashed lines) 
as a function of $k$ in the multi-$k$ approach are shown in Figure~\ref{fig:stream}.
The behaviour of AP and AL 
is that expected for the four experimental setups defined above, but 
the conventional DAL measure ($s=1.0$) abruptly suffers the effect 
of not being able to recover from 
accumulated delays due to the cost of write operations. In contrast, 
DAL with $s=0.95$ exhibits a smooth interpretable behaviour as a result 
of compensating for re-segmentation errors.
Moreover, the gap
between "In. Seg." and "In. Seg. + Out. Seg." is not significant, therefore we believe that, if the
translation quality is good enough, the automatic re-segmentation
process is an acceptable way of computing stream-level
latencies. Lastly, as expected, if we use an oracle system that
outputs the reference translation with the appropriate writing rate
for each sentence ("Policy + In. Seg. + Out. Seg."), the obtained AL and DAL values
are very close to the theoretical value $k$. If we compute DAL using $s=0.95$, we
obtain similar values without the need of using any oracle, while
accounting for the additional cost of write operations.

Thus, unlike
the Concat-1 approach, our stream-level approach 
is highly effective for providing 
interpretable and accurate latency measures.

\section{Conclusions}
\label{sec:con}

In this work, an adaptation of the current latency measures to a
streaming setup is proposed motivated by the lack of interpretability 
of sentence-level latency measures in this setup.

This adaptation basically consists in the modification of the 
conventional latency measures to move from a sentence-level evaluation 
based on a local delay function to a
stream-level estimation by using a global delay function that
keeps track of delays across the whole translation process. 
At the same time,
a re-segmentation approach has been proposed to compute these latency measures on any arbitrary segmentation of the input stream used by
the translation model. The resulting measures are highly interpretable 
and coherent accounting for the actual behaviour of the simultaneous 
translation system in a real streaming scenario.

\section*{Acknowledgements}
The research leading to these results has received funding from the
European Union’s Horizon 2020 research and innovation program under
grant agreement no. 761758 (X5Gon) and 952215 (TAILOR) and Erasmus+
Education program under grant agreement no. 20-226-093604-SCH; the
Government of Spain’s research project Multisub, ref.
RTI2018-094879-B-I00 (MCIU/AEI/FEDER,EU) and FPU scholarships
FPU18/04135; and the Generalitat Valenciana’s research project
Classroom Activity Recognition, ref. PROMETEO/2019/111.

\bibliography{anthology,custom}
\bibliographystyle{acl_natbib}

\appendix

\section{Reproducibility of proposed measures}
The code for the proposed latency measures, as well as all the translations have been published \footnote{\url{https://github.com/jairsan/Stream-level_Latency_Evaluation_for_Simultaneous_Machine_Translation}}. A script is included
to reproduce the results reported in the paper. The full results for the Concat-1 method are reported on Tables \ref{tab:concat_1_al_app} and \ref{tab:concat_1_dal_app}

\begin{table*}
\centering \caption{Stream-level AL as a function of $k$, computed using the Concat-1 approach on the IWSLT 2010 German-English dev set. \label{tab:concat_1_al_app}}
\begin{tabular}{l||rrrrrrrrrr}
System & 1 & 2 & 3 & 4 & 5 & 6 & 7 & 8 & 9 & 10 \\ \hline
Real & -9.7 & -12.0 & -45.2 & -23.7 & -8.5 & -4.4 & -17.4 & -13.6 & -14.2 & -12.2 \\
~ +In-Seg Oracle & -42.9 & -29.0 & 17.4 & -10.1 & 25.5 & 3.8 & 9.7 & 5.3 & 2.7 & 4.7 \\
~~~  + Policy Oracle & 14.2 & 15.1 & 16.0 & 16.8 & 17.6 & 18.2 & 18.9 & 19.5 &  20.1 & 20.6 \\
\end{tabular}
\end{table*}

\begin{table*}
\centering \caption{Stream-level DAL as a function of $k$, computed using the Concat-1 approach on the IWSLT 2010 German-English dev set. \label{tab:concat_1_dal_app}}
\begin{tabular}{l||rrrrrrrrrr}
System & 1 & 2 & 3 & 4 & 5 & 6 & 7 & 8 & 9 & 10 \\ \hline
Real &  15.0 & 11.0 & 17.4 & 11.3 & 20.3 & 25.1 & 11.3 & 14.4 & 17.7 & 17.9 \\
~ +In-Seg Oracle & 4.5 & 8.5 & 37.1 & 24.6 &  52.3 & 27.8 & 21.5 & 33.9 & 31.2 & 31.8 \\
~~~  + Policy Oracle & 85.8 & 86.7 & 87.7 & 88.7 & 89.7 & 90.7 & 91.7 & 92.7 & 93.6 & 94.6 \\
\end{tabular}
\end{table*}

\section{MT System}

\begin{table}
\centering
 \caption{Corpus used for MT model training \label{ap:mt_corpus}}
\small 
	\begin{tabular}{l||rrr}
& & \multicolumn{2}{c}{tokens(M)} \\ 
Corpus & sentences(M) & German & English \\ 
\hline
News Commentary  &  0.3  &  7.4  &  7.2  \\
WikiTitles  &  1.3  &  2.7  &  3.1  \\ 
Europarl  &  1.8  &  42.5  &  45.5  \\ 
Rapid  &  1.5  &  26.0  &  26.9  \\ 
MuST-C  &  0.2  &  3.9  &  4.2  \\ 
Ted  &  0.2  &  3.3  &  3.6  \\ 
LibriVox  &  0.1  &  0.9  &  1.1  \\ 
Paracrawl  &  31.4  &  465.2  &  502.9  \\
	\end{tabular}
\end{table}

Table \ref{ap:mt_corpus} lists the corpus that were selected for training out of the IWSLT 2020 allowed data \footnote{\url{http://iwslt2020.ira.uka.de/doku.php?id=offline_speech_translation}}.

The multi-$k$ system has been trained with the official implementation \footnote{\url{https://github.com/elbayadm/attn2d}}. The model was trained for 0.5M steps on a machine with 2 2080Ti GPUs, which took 6 days. The following command was used to train it:
{ 
\tiny
\begin{verbatim}
fairseq-train $CORPUS_FOLDER \
-s $SOURCE_LANG_SUFFIX \
-t $TARGET_LANG_SUFFIX \
--user-dir $FAIRSEQ/examples/waitk \
--arch waitk_transformer_base \
--share-decoder-input-output-embed \
--left-pad-source False \
--multi-waitk \
--optimizer adam \
--adam-betas '(0.9, 0.98)' \
--clip-norm 0.0 \
--lr-scheduler inverse_sqrt \
--warmup-init-lr 1e-07 \
--warmup-updates 4000 \
--lr 0.0005 \
--min-lr 1e-09 \
--dropout 0.3 \
--weight-decay 0.0 \
--criterion label_smoothed_cross_entropy \
--label-smoothing 0.1 \
--max-tokens 4000 \
--update-freq 4 \
--save-dir $MODEL_OUTPUT_FOLDER \
--no-progress-bar \
--log-interval 100 \
--max-update 500000 \
--save-interval-updates 10000 \
--keep-interval-updates 20 \
--ddp-backend=no_c10d \
--fp16

\end{verbatim}
}

\section{Segmenter System}
The Direct Segmentation system has been trained with the official implementation \footnote{\url{https://github.com/jairsan/Speech_Translation_Segmenter}}. The ted corpus was used as training data (See Table \ref{ap:mt_corpus}). The following command was used to train the segmenter system:
{ \tiny
\begin{verbatim}
len=11
window=0
python3 train_text_model.py \
--train_corpus train.ML$len.WS$window.txt \
--dev_corpus  dev.ML$len.WS$window.txt \
--output_folder $output_folder \
--vocabulary $corpus_folder/train.vocab.txt \
--checkpoint_interval 1 \
--epochs 15 \
--rnn_layer_size 256 \
--embedding_size 256 \
--n_classes 2 \
--batch_size 256 \
--min_split_samples_batch_ratio 0.3 \
--optimizer adam \
--lr 0.0001 \
--lr_schedule reduce_on_plateau \
--lr_reduce_patience 5 \
--dropout 0.3 \
--model_architecture ff-text \
--feedforward_layers 2 \
--feedforward_size 128 \
--sample_max_len $len \
--sample_window_size $window
\end{verbatim}
}

\end{document}